\pdfoutput=1
\documentclass[11pt,a4paper]{article}
\usepackage[hyperref]{acl2021}
\usepackage{times}
\usepackage{latexsym}
\usepackage{amsmath}
\usepackage{multirow}
\usepackage{booktabs}
\usepackage{graphicx}
\usepackage{amsfonts}
\usepackage{subcaption}

\usepackage{microtype}

\aclfinalcopy 
\def\aclpaperid{527} 

\title{Out-of-Scope Intent Detection with Self-Supervision and\\ Discriminative Training}

\author{
Li-Ming Zhan$^1$ \quad Haowen Liang$^1$\thanks{~ Equal contribution.}  \quad Bo Liu$^{1*}$ \quad Lu Fan$^1$ \quad Xiao-Ming Wu$^1$\Thanks{~ Corresponding author.} \\
\quad \bf Albert Y.S. Lam$^2$ \\ 
Department of Computing, The Hong Kong Polytechnic University, Hong Kong S.A.R.$^1$ \\
Fano Labs, Hong Kong S.A.R.$^2$ \\
{\tt \{lmzhan.zhan, michael.liang, doc-bo.liu\}@connect.polyu.edu.hk}\\
{\tt \{cslfan, csxmwu\}@comp.polyu.edu.hk, albert@fano.ai}\\
}
\date{}

\begin{document}
\maketitle
\begin{abstract}

Out-of-scope intent detection is of practical importance in task-oriented dialogue systems. Since the distribution of outlier utterances is arbitrary and unknown in the training stage, existing methods commonly rely on strong assumptions on data distribution such as mixture of Gaussians to make inference, resulting in either complex multi-step training procedures or hand-crafted rules such as confidence threshold selection for outlier detection.
In this paper, we propose a simple yet effective method to train an out-of-scope intent classifier in a fully end-to-end manner by simulating the test scenario in training, which requires no assumption on data distribution and no additional post-processing or threshold setting. Specifically, we construct a set of pseudo outliers in the training stage, by generating synthetic outliers using inliner features via self-supervision and sampling out-of-scope sentences from easily available open-domain datasets. The pseudo outliers are used to train a discriminative classifier that can be directly applied to and generalize well on the test task. We evaluate our method extensively on four benchmark dialogue datasets and observe significant improvements over state-of-the-art approaches.
Our code has been released at \url{https://github.com/liam0949/DCLOOS}.  


\begin{figure}
\begin{subfigure}{0.49\textwidth}
  \centering
  \includegraphics[width=.97\linewidth,height=0.152\textheight]{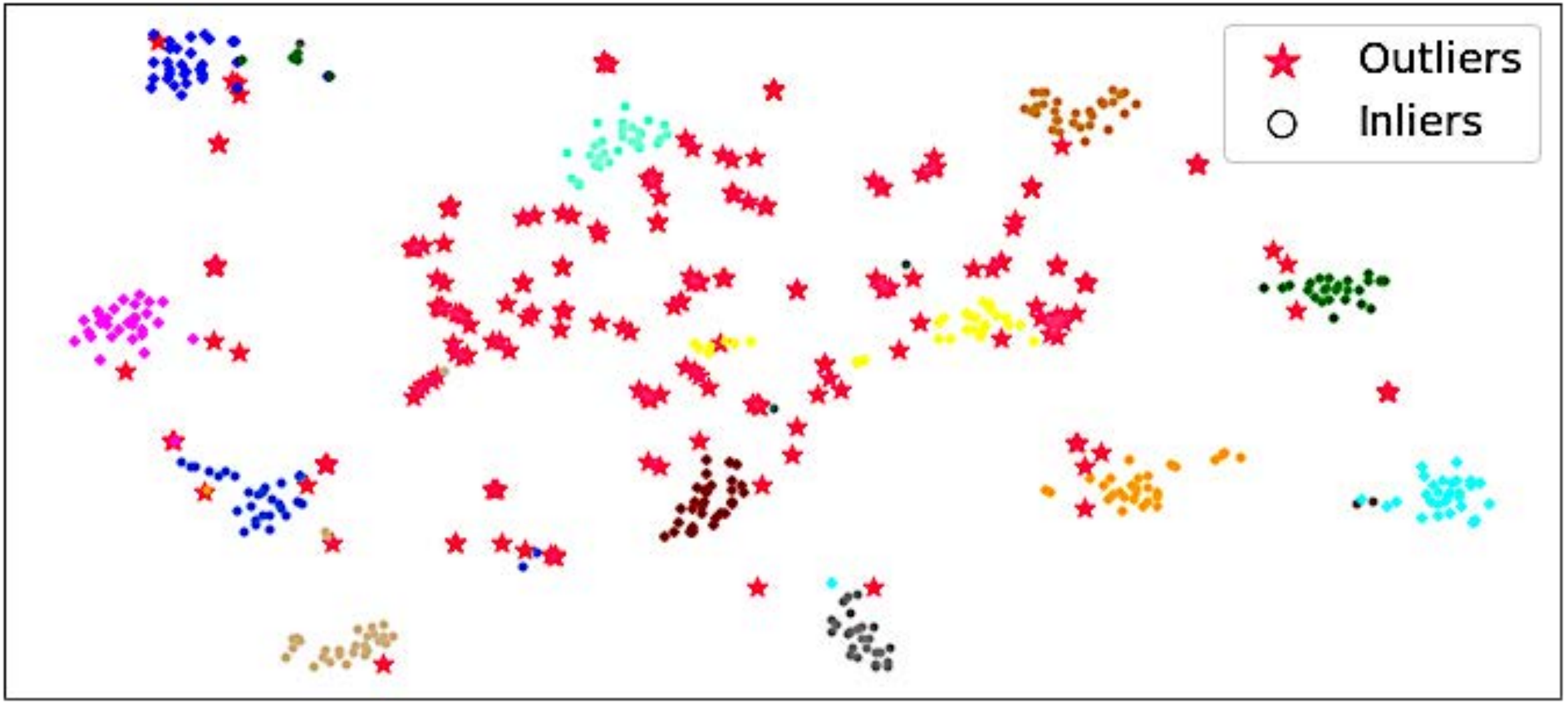}
\end{subfigure}
\begin{subfigure}{0.49\textwidth}
  \centering
  \includegraphics[width=.97\linewidth,height=0.152\textheight]{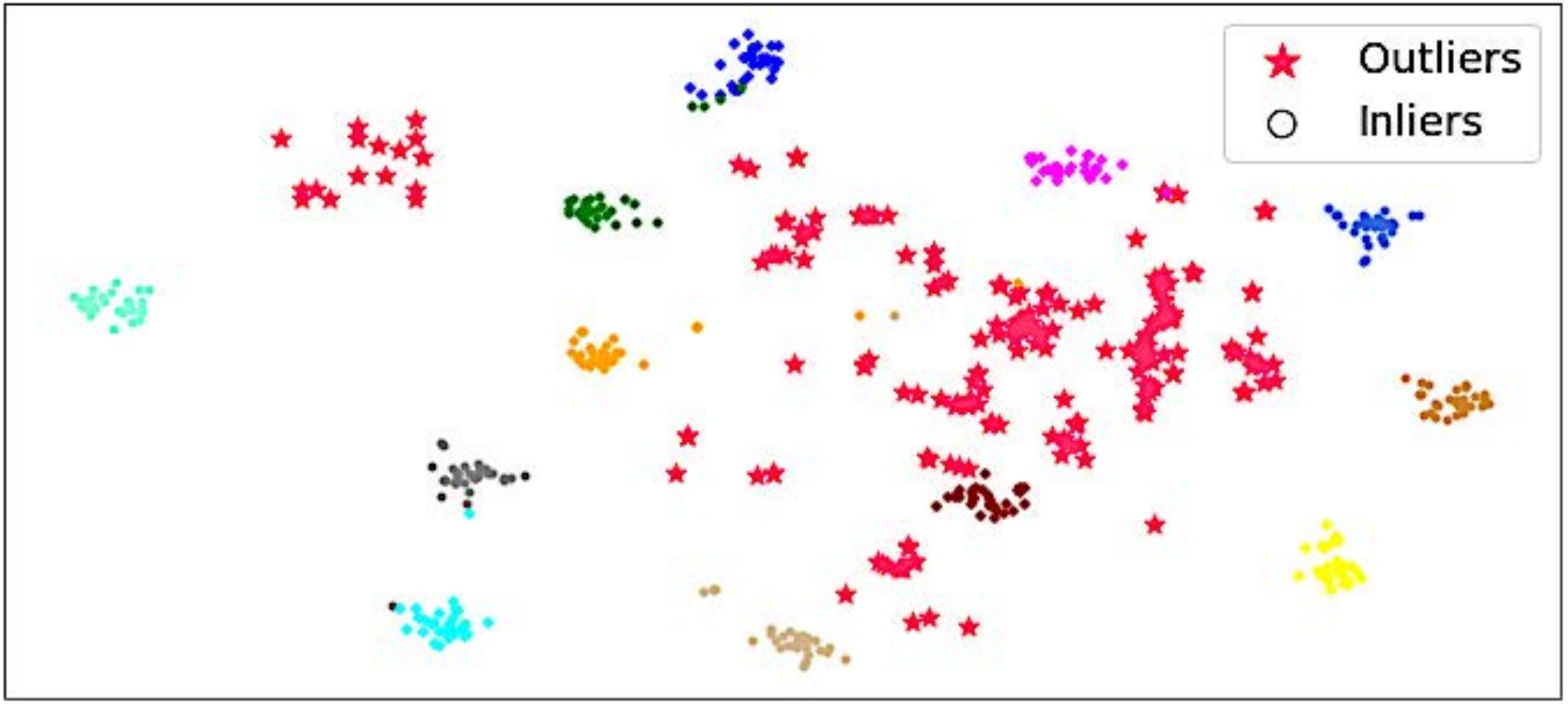}
\end{subfigure}
\vspace{-5pt}
\caption{t-SNE visualization of the learned embeddings of the test samples of CLINC150. Top: Previous $K$-way training; Bottom: Our proposed $(K+1)$-way training. Better view in color and enlarged.}
\label{fig:tsne}
\end{figure}
\end{abstract}

\section{Introduction}

Conversational system is becoming an indispensable component in a variety of AI applications and acts as an interactive interface provided to users to improve user experience. Language understanding is essential for conversational systems to provide appropriate responses to users, and intent detection is usually the first step of language understanding. The primary goal is to identify diverse intentions behind user utterances, which is often formalized as a classification task. However, intent classes defined during training are inevitably inadequate to cover all possible user intents at the test stage due to the diversity and randomness of user utterances. Hence, out-of-scope (or unknown) intent detection is essential, which aims to develop a model that can accurately identify known (seen in training) intent classes while detecting the out-of-scope classes that are not encountered during training.

Due to the practical importance of out-of-scope intent detection, recent efforts have attempted to solve this problem by developing effective intent classification models. In general, previous works approach this problem by learning \emph{decision boundaries} for known intents and then using some confidence measure to distinguish known and unknown intents. For examples, LMCL~\citep{DBLP:conf/acl/LinX19} learns the decision boundaries with a margin-based optimization objective, and SEG~\citep{yan-etal-2020-unknown} assumes the known intent classes follow the distribution of mixture of Gaussians. After learning the decision boundaries, an off-the-shell outlier detection algorithm such as LOF~\citep{10.1145/335191.335388} is commonly employed to derive confidence scores  \cite{yan-etal-2020-unknown,DBLP:conf/emnlp/ShuXL17, DBLP:conf/acl/LinX19, DBLP:conf/iclr/HendrycksG17}.
If the confidence score of a test sample is lower than a predefined threshold, it is identified as an outlier. 

However, it may be problematic to learn decision boundaries solely based on the training examples of known intent classes. First, if there are sufficient training examples, the learned decision boundaries can be expected to generalize well on known intent classes, but not on the unknown. Therefore, extra steps are required in previous methods, such as using an additional outlier detection algorithm at the test stage or adjusting the confidence threshold by cross-validation. On the other hand, if there are not sufficient training examples, the learned boundaries may not generalize well on both known and unknown intents. As a result, these methods often underperform when not enough training data is given. Hence, it is important to provide learning signals of unknown intents at the training stage to overcome these limitations.



In contrast to previous works, we adopt a different approach by explicitly modeling the distribution of unknown intents. Particularly, we construct a set of pseudo out-of-scope examples to aid the training process. We hypothesize that in the semantic feature space, real-world outliers can be well represented in two types: ``\emph{hard}'' outliers that are geometrically close to the inliers and ``\emph{easy}'' outliers that are distant from the inliners. For the ``hard'' ones, we construct them in a self-supervised manner by forming convex combination of the features of inliers from different classes. 
For the ``easy'' ones, the assumption is that they are very unrelated to the known intent classes, so they can be used to simulate the randomness and diversity of user utterances. They can be easily constructed using public datasets. For example, in our experiments, we randomly collect sentences from datasets of other NLP tasks such as question answering and sentiment analysis as open-domain outliers. 

In effect, by constructing pseudo outliers for the unknown class during training, we form a consistent $(K+1)$ classification task ($K$ known classes + 1 unknown class) for both training and test. Our model can be trained with a cross-entropy loss and directly applied to test data for intent classification and outlier detection without requiring any further steps.
As shown in Figure~\ref{fig:tsne} (better view in color and enlarged), our method can learn better utterance representations, which make each known intent class more compact and push the outliers away from the inliers. Our main contributions are summarized as follows.
\begin{itemize}

\item We propose a novel out-of-scope intent detection approach by matching training and test tasks to bridge the gap between fitting to training data and generalizing to test data.

\item We propose to efficiently construct two types of pseudo outliers by using a simple self-supervised method and leveraging publicly available auxiliary datasets.

\item We conduct extensive experiments on four real-world dialogue datasets to demonstrate the effectiveness of our method and perform a detailed ablation study.
\end{itemize}

\section{Related Work}
\subsection{Out-of-Distribution Detection}
Early studies on outlier detection often adopt unsupervised clustering methods to detect malformed data~\citep{DBLP:journals/air/HodgeA04, chandola2009anomaly,zimek2012survey}. In recent years, a substantial body of work has been directed towards improving the generalization capacity of machine learning models on out-of-distribution (OOD) data~\citep{DBLP:journals/pieee/RuffKVMSKDM21, hendrycks2020many}. 
\citet{DBLP:conf/iclr/HendrycksG17} find that simple statistics derived from the outputting softmax probabilities of deep neural networks can be helpful for detecting OOD samples. Following this work, \citet{dcbe7abf4db64d1b89bf9802585660ed} propose to use temperature scaling and add small perturbation to input images to enlarge the gap between in-scope and OOD samples. 
\citet{lee2017training} propose to add a Kullback-Leibler divergence term in the loss function to encourage assigning lower maximum scores to OOD data. 

Recently, there is a line of work that employs synthetic or real-world auxiliary datasets to provide learning signals for improving model robustness under various forms of distribution shift~\citep{DBLP:journals/corr/GoodfellowSS14,DBLP:journals/corr/abs-1907-07640,DBLP:conf/icml/HendrycksLM19,lee2017training}. Particularly, \citet{hendrycks2018deep} propose to leverage large-scale public datasets to represent outliers during training time and form a regularization term based on that. This idea is similar to our proposal of constructing open-domain outliers, but we use a simpler, end-to-end, $(K+1)$-way discriminative training procedure without any regularization term or threshold parameter.  

\begin{figure*}[t]
    \centering
    \includegraphics[width=150mm]{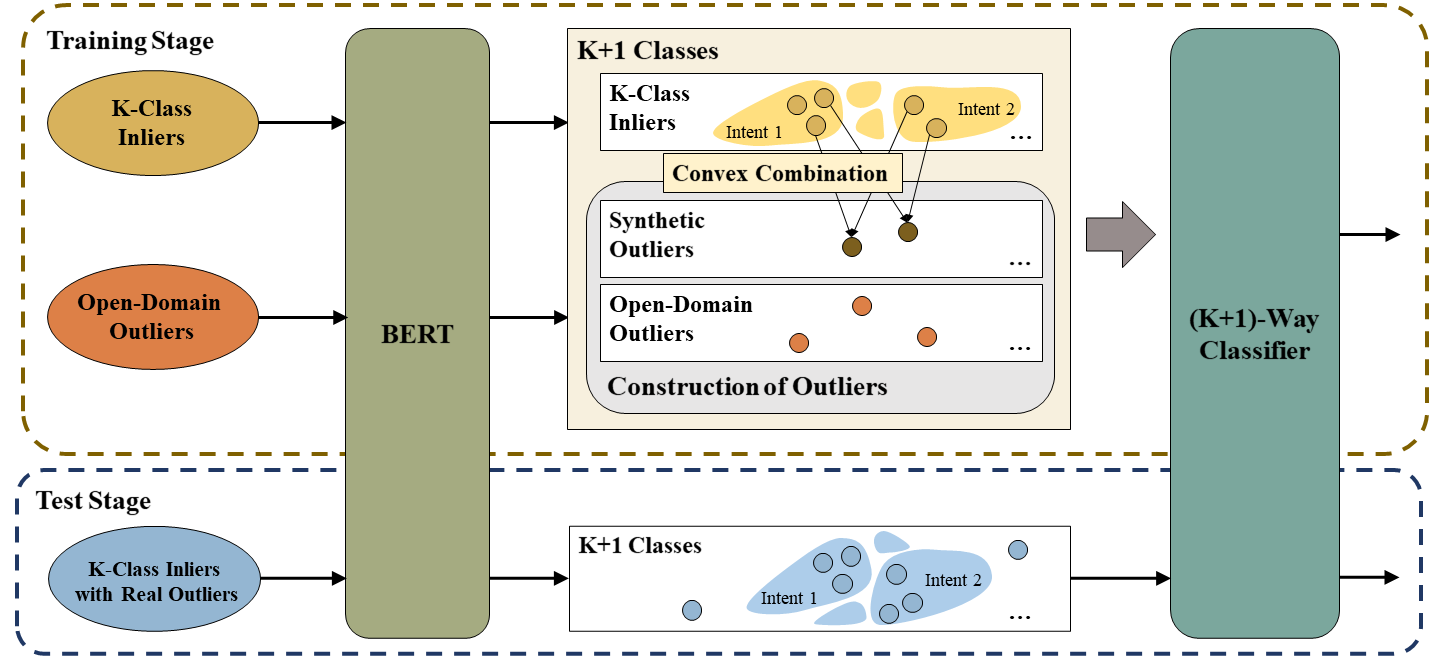}
    \caption{An illustration of our proposed method. We use BERT as the utterance encoder. At training stage, we train a (K+1)-way classifier by constructing two types of pseudo outliers. The open-domain outliers are collected from an auxiliary dataset disjoint from both the training and test data. The synthetic self-supervised outliers are generated during training by random convex combinations of features of inliers from different known classes.}
    \label{fig:framework}
\end{figure*}
\subsection{Out-of-Scope Intent Detection}
While \citet{DBLP:conf/acl/HendrycksLWDKS20} find that pretrained transformer-based models like BERT are intrinsically more robust to OOD data, they suggest that there are still margins for improvement. Therefore, we build our model on top of BERT to improve intent detection under significant distribution shift. Previous methods for out-of-scope (or out-of-distribution) intent detection are commonly threshold-based, where models output a decision score and then compare it with a threshold that is predefined or selected by cross-validation. 

There are mainly three branches of related work. The first group uses a confidence score which determines the likelihood of an utterance being out-of-scope. 
For example, \citet{DBLP:conf/emnlp/ShuXL17} build $m$ binary Sigmoid classifiers for $m$ known classes respectively and select a threshold to reject OOD inputs that may have lower probabilities than the threshold across all $m$ classifiers. 
Similar to the OOD data generation method used in \citet{lee2017training}, \citet{ryu-etal-2018-domain} employ GAN~\citep{goodfellow2014generative} to generate simulated OOD examples with the generator and learn to reject simulated OOD examples with the discriminator.

The second group identifies out-of-scope sentences through reconstruction loss. For example, \citet{10.1016/j.patrec.2017.01.008} build an autoencoder to encode and decode in-scope utterances and obtain reconstruction loss by comparing input embeddings with decoded ones. Out-of-scope utterances result in higher reconstruction loss. 

The third group leverages off-the-shell outlier detection algorithms such as local outlier factor (LOF)~\citep{10.1145/335191.335388}, one-class SVM~\citep{6790022}, robust covariance estimators~\citep{article}, and isolation forest~\citep{10.1109/ICDM.2008.17} to detect out-of-scope examples. Utterance embeddings belonging to a specific class will be mapped to the corresponding cluster (usually modeled by a Gaussian distribution) while out-of-scope samples will be pushed away from all in-scope clusters. Examples of this kind include SEG~\citep{DBLP:conf/acl/YanFLLZWL20} and LMCL~\citep{DBLP:conf/acl/LinX19}. Very recently, \citet{Zhang_Xu_Lin_2021} propose to learn adaptive decision boundaries after pre-training instead of using off-the-shell outlier detection algorithms.

In addition, some other work focuses on out-of-scope detection in few-shot scenarios. \citet{tan-etal-2019-domain} leverage independent source datasets as simulated OOD examples to form a hinge loss term. \citet{zhang-etal-2020-discriminative} propose to pretrain BERT by a natual language understanding task with large-scale training data to transfer useful information for few-shot intent detection. 

Finally, for our proposal of constructing synthetic outliers, the most similar method is {Mixup} proposed by \citet{DBLP:conf/iclr/ZhangCDL18}. However, their method is designed for data augmentation to enhance in-distribution performance and requires corresponding combinations in the label space \cite{DBLP:conf/nips/ThulasidasanCBB19}.

\section{Methodology}

\paragraph{Problem Statement}
In a dialogue system, given $K$ predefined intent classes $S_{\text{known}}=\{C_i\}_{i=1}^K$, an unknown intent detection model aims at predicting the category of an utterance $u$, which may be one of the known intents or an out-of-scope intent $C_{\text{oos}}$. Essentially, it is a $K+1$ classification problem at the test stage. At the training stage, a set of $N$ labeled utterances $\mathcal{D}_{l} = \{ (x_i,c_i) \mid c_i\in S_{\text{known}})\}_{i=1}^{N}$ is provided for training. Previous methods typically train a $K$-way classifier for the known intents. 

\paragraph{Overview of Our Approach} 
The mismatch between the training and test tasks, i.e., $K$-way classification vs. $(K+1)$-way classification, leads to the use of strong assumptions and additional complexity in previous methods. Inspired by recent practice in meta learning to simulate test conditions in training~\citep{vinyals2016matching}, we propose to match the training and test settings. In essence, as shown in Figure~\ref{fig:framework}, we formalize a $(K+1)$-way classification task in the training stage by constructing out-of-scope samples via self-supervision and from open-domain data. Our method simply trains a $(K+1)$-way classifier without making any assumption on the data distribution. After training, the classifier can be readily applied to the test task without any adaptation or post-processing.
In the following, we elaborate on the details of our proposed method, including representation learning, construction of pseudo outliers, and discriminative training.


\subsection{Representation Learning}
We employ BERT~\citep{DBLP:conf/naacl/DevlinCLT19} -- a deep Transformer network as text encoder. Specifically, we take the $d$-dimensional output vector of the special classification token [CLS] as the representation of an utterance $u$, i.e.,
$$h = \text{BERT}(u) \in \mathbb{R}^d,$$ 
where $d=768$ by default. The training set $\mathcal{D}_{l}$ is then mapped to $\mathcal{D}_{l}^{tr} = \{(h_i, c_i)\mid h_i = \text{BERT}(u_i), (u_i, c_i)\in \mathcal{D}_{l}\}_{i=1}^N$ in the feature space.




\subsection{Construction of Outliers}\label{subsec: building outlier}


We construct two different types of pseudo outliers to be used in the training stage: synthetic outliers that are generated by self-supervision, and open-domain outliers that can be easily acquired. 




\paragraph{Synthetic Outliers by Self-Supervision} To improve the generalization ability of the unknown intent detection model, we propose to generate ``\emph{hard}'' outliers in the feature space, which may have similar representations to the inliers of known intent classes. We hypothesize that those outliers may be geometrically \emph{close} to the inliers in the feature space. Based on this assumption, we propose a self-supervised method to generate the ``hard'' outliers using the training set $\mathcal{D}^{tr}_{l}$.  

Specifically, in the feature space, we generate synthetic outliers by using convex combinations of the features of  inliers from different intent classes:
\begin{flalign}
\begin{aligned}
	h^{oos} =\theta * h_{\beta} + (1-\theta)*h_{\alpha},
\end{aligned}
\end{flalign}
where $h_{\beta}$ and $h_{\alpha}$ are the representations of two utterances which are randomly sampled from different intent classes in $\mathcal{D}_{l}^{tr}$, i.e., $c_{\beta} \neq c_{\alpha}$, and $h^{oos}$ is the synthetic outlier. For example, $\theta$ can be sampled from a uniform distribution $U(0, 1)$. In this case, when $\theta$ is close to $0$ or $1$, it will generate ``harder'' outliers that only contain a small proportion of mix-up from different classes. In essence, ``hard'' outliers act like support vectors in SVM~\citep{cortes1995support}, and ``harder'' outliers could help to train a more discriminative classifier. 


The generated outliers $h^{oos}$ are assigned to the class of $C_{oos}$, the $(K+1)$-th class in the feature space, forming a training set
\begin{flalign}
\begin{aligned}
	\mathcal{D}_{co}^{tr} = \{(h_{i}^{oos}, c_i = C_{oos}) \}_{i=1}^{M}.
\end{aligned}
\end{flalign}
Notice that since the outliers are generated in the feature space, it is very efficient to construct a large outlier set $\mathcal{D}_{co}^{tr}$.



\paragraph{Open-Domain Outliers} In practical dialogue systems, user input can be arbitrary free-form sentences. To simulate real-world outliers and provide learning signals representing them in training, we propose to construct a set of open-domain outliers, which can be easily obtained. Specifically, the set of free-form outliers $\mathcal{D}_{fo}$ can be constructed by collecting sentences from various public datasets that are disjoint from the training and test tasks. There are many datasets available, including the question answering dataset SQuaD 2.0~\citep{DBLP:conf/acl/RajpurkarJL18}, the sentiment analysis datasets Yelp~\citep{DBLP:conf/cikm/MengSZ018} and IMDB~\citep{DBLP:conf/acl/MaasDPHNP11}, and dialogue datasets from different domains. 

In the feature space, $\mathcal{D}_{fo}$ is mapped to
$\mathcal{D}_{fo}^{tr} = \{(h^{oos}_i, c_i=C_{oos})\mid h^{oos}_i=\text{BERT}(u_i), u_i \in \mathcal{D}_{fo}\}_{i=1}^H$.

Both synthetic outliers and open-domain outliers are easy to construct. As will be demonstrated in Section~\ref{sec:exp}, both of them are useful, but synthetic outliers are much more effective than open-domain outliers in improving the generalization ability of the trained $(K+1)$-way intent classifier. 


\subsection{Discriminative Training}

After constructing the pseudo outliers, in the feature space, our training set $\mathcal{D}^{tr}$ now consists of a set of inliers $\mathcal{D}_{l}^{tr}$ and two sets of outliers $\mathcal{D}_{co}^{tr}$ and $\mathcal{D}_{fo}^{tr}$, i.e., $\mathcal{D}^{tr} = \mathcal{D}_{l}^{tr} \cup \mathcal{D}_{co}^{tr} \cup \mathcal{D}_{fo}^{tr}$ and $|\mathcal{D}^{tr}| = N+M+H$. Therefore, in the training stage, we can train a $(K+1)$-way classifier  with the intent label set $S= S_{known} \cup \{C_{oos}\}$, which can be directly applied in the test stage to identify unknown intent and classify known ones. In particular, we use a multilayer perceptron network, $\Phi(\cdot)$, as the classifier in the feature space. The selection of the classifier is flexible, and the only requirement is that it is differentiable. Then, we train our model using a cross-entropy loss:
\begin{align*}
	\mathcal{L} = 
	- \frac{1}{|\mathcal{D}^{tr}|}\sum_{\mathcal{D}^{tr}}\log \frac{\exp(\Phi(h_i)^{c_i}/\tau)}{\sum_{j\in S}\exp(\Phi(h_i)^j/\tau)},
\end{align*}
where $\Phi(h_i)^{c_i}$ refers to the output logit of $\Phi(\cdot)$ for the ground-truth class $c_i$, and $\tau \in \mathbb{R}^+$ is an adjustable scalar temperature parameter.

\section{Experiments}\label{sec:exp}
\begin{table*}[ht!]
\centering
\scalebox{0.85}{
\begin{tabular}{cc|cc|cc|cc|cc}
\specialrule{0.1em}{0.2em}{0.2em}
\multicolumn{2}{c}{}&\multicolumn{2}{c}{CLINC150}&\multicolumn{2}{c}{StackOverflow}&\multicolumn{2}{c}{Banking}&\multicolumn{2}{c}{M-CID-EN}\\
&Methods & Accuracy & Macro-F1& Accuracy & Macro-F1& Accuracy & Macro-F1& Accuracy & Macro-F1\\
\specialrule{0.05em}{0.1em}{0.2em}
\multirow{6}{*}{25\%} 
                    &MSP&66.60 & 51.20 &33.94 &45.68  &48.15 &48.47 &52.05&43.14   \\
                    &DOC&64.43 & 44.60 &60.68 &60.51  &37.78 &46.35 &49.32&46.59  \\
                    &SEG&72.86 & 65.44 &47.00 &52.83  &51.11 &55.68 &44.51&50.14 \\
                    &LMCL& 68.57&62.42 &41.60 &48.21  &52.77 &56.73 &41.44&46.99  \\
                    &Softmax&76.50& 67.74 &46.17 &50.78  &57.88 & 58.32&41.95&45.46   \\
                    &\bf {Ours} &\bf88.44 & \bf 80.73 &\bf68.74 & \bf 65.64 & \bf 74.11&\bf 69.93&\bf 87.08&\bf 79.67\\
\specialrule{0.05em}{0.1em}{0.2em}
\multirow{6}{*}{50\%} 
                    &MSP& 68.61& 51.20 &56.33 &62.92  &53.83 &65.33 &61.21&54.33  \\
                    &DOC& 62.46 & 70.01  &61.62 &68.97  &58.29 &57.30  &59.97&62.28 \\
                    &SEG& 77.05 & 79.42 & 68.50 & 74.18 &68.44 &76.48  &67.91&72.37 \\
                    &LMCL& 78.63& 80.42 &64.34 &71.80 & 63.59 &73.99 &63.42&69.04  \\
                    &Softmax&82.47&82.86 &65.96 &71.94  &67.44 &74.19 &64.72&69.35 \\
                    &\bf {Ours}&\bf 88.33 &\bf 86.67 &\bf 75.08 &\bf 78.55 & \bf 72.69&\bf 79.21&\bf81.05&\bf79.73\\
\specialrule{0.05em}{0.1em}{0.2em}
\multirow{6}{*}{75\%} 
                    &MSP&73.41 &81.81  &76.73 & 77.63 &71.92 &80.77 &72.89&77.34  \\
                    &DOC&74.63 & 78.63 & 63.98 &62.07  &72.02 &78.04 &69.79&71.18  \\
                    &SEG&81.92 & 86.57  &80.83 &84.78  &78.87 &85.66 &75.73&79.97 \\
                    &LMCL& 84.59&88.21 &80.02 & 84.47  &78.66 &85.33 &77.11&80.96  \\
                    &Softmax& 86.26& 89.01 &77.41 &82.28  &78.20 &84.31 &76.99& 80.82 \\
                    &\bf {Ours}&\bf88.08 & \bf89.43 &\bf81.71 & \bf85.85 &\bf81.07 &\bf86.98&\bf80.24&\bf82.75\\
\specialrule{0.1em}{0.1em}{0.2em}
\end{tabular}
}
\caption{\label{mainresult1}
Overall accuracy and macro f1-score for unknown intent detection with different proportion of seen classes. For each setting, the best result is marked in bold. 
}
\end{table*}
In this section, we present the experimental results of our proposed method on the targeted task of unknown intent detection. Given a test set comprised of known and unknown intent classes, the primary goal of an unknown intent detection model is to assign correct intent labels to utterances in the test set. Notice that the unknown intent label $C_{oos}$ is also included as a special class for prediction. 
\begin{table}
\centering
\scalebox{0.76}{
\begin{tabular}{l|c|c|c|c}
\specialrule{0.05em}{0.1em}{0.2em}
Dataset & Vocab & Avg. Length & Samples & Classes\\
\specialrule{0.05em}{0.1em}{0.2em}
CLINC150&8,376&8.31&23,700&150\\
StackOverflow&17,182&9.18&20,000&20\\
Banking&5028&11.9&13,083&77\\
M-CID-EN&1,254&6.74&1,745&16\\
\specialrule{0.05em}{0.1em}{0.2em}
\end{tabular}
}
\caption{\label{table:datsets_stats}
Dataset statistics.
}
\end{table}

\begin{table*}[t]
\centering
\scalebox{0.90}{
\begin{tabular}{cc|cc|cc|cc|cc}
\specialrule{0.1em}{0.2em}{0.2em}
\multicolumn{2}{c}{}&\multicolumn{2}{c}{CLINC150}&\multicolumn{2}{c}{StackOverflow}&\multicolumn{2}{c}{Banking}&\multicolumn{2}{c}{M-CID-EN}\\
&Methods & Unknown & Known& Unknown & Known& Unknown & Known& Unknown & Known\\
\specialrule{0.05em}{0.1em}{0.2em}
\multirow{6}{*}{25\%} 
                    &MSP&73.20 &50.62  &22.59 &50.30  &49.98 &48.39 &56.27&37.86  \\
                    &DOC&71.08 &43.91  &66.11 &59.39  &31.41 &47.14  &53.08&44.92 \\
                    &SEG&79.90 &65.06  &46.17 &54.16  &53.22 &55.81 &42.73&51.99  \\
                    &LMCL&75.61 &62.01  &38.85 &50.15 &55.29 &56.81  &36.99&49.50 \\
                    &Softmax&83.04 &67.34  &45.52 &51.83  &62.52 &58.10 &35.39&46.22  \\
                    &\bf {Ours} &\bf92.35 & \bf80.43 &\bf74.86 & \bf63.80 & \bf80.12&\bf69.39&\bf91.15&\bf76.80\\
\specialrule{0.05em}{0.1em}{0.2em}
\multirow{6}{*}{50\%} 
                    &MSP&57.78 &68.03  &35.18 &70.09  &29.31 &66.28 &58.55&53.80  \\
                    &DOC&57.62 &70.17  &47.96 &71.07  &49.88 &57.50  &47.22&64.16 \\
                    &SEG&78.02 &79.43  &60.89 &75.51 &60.42 &76.90&61.04 &73.80\\
                    &LMCL&79.89 &80.42  &53.12 &71.80  &50.30 &74.62  &51.11&71.29 \\
                    &Softmax&84.19 &82.84  &56.80 &73.45  &60.28 &74.56 &56.30&70.98  \\
                    &\bf {Ours}&\bf90.30 &\bf86.54 &\bf71.88 &\bf79.22 & \bf67.26&\bf79.52&\bf82.44&\bf79.39\\
\specialrule{0.05em}{0.1em}{0.2em}
\multirow{6}{*}{75\%} 
                    &MSP&57.83 &82.02  &41.73 &80.03  &23.86 &81.75 &39.56&80.50  \\
                    &DOC&64.62 &78.76  &49.50&62.91  &39.47 &78.72 &49.41&72.99  \\
                    &SEG&76.12 &86.67  &62.30 &86.28  &54.43 &86.20 &51.51&82.34  \\
                    &LMCL&80.42 &88.28  &61.40 &84.47  &53.26 &85.89  &54.61&83.16 \\
                    &Softmax&83.12 &\bf89.61  &54.07&84.11  &56.90 &84.78 &58.73&82.66  \\
                    &\bf {Ours}&\bf86.28 &89.46 &\bf65.44 & \bf87.22 &\bf60.71 &\bf87.47&\bf69.00&\bf83.89\\
\specialrule{0.1em}{0.1em}{0.2em}
\end{tabular}
}
\caption{\label{mainresult2}
Macro f1-score of the known classes and f1-score of the unknown class with different proportion of seen classes. For each setting, the best result is marked in bold.
}
\end{table*}
\subsection{Datasets and Baselines}

We evaluate our proposed method on four benchmark datasets as follows, three of which are newly released dialogue datasets designed for intent detection. The statistics of the datasets are summarized in Table~\ref{table:datsets_stats}. 

\textbf{CLINC150}~\citep{larson2019evaluation} is a dataset specially designed for out-of-scope intent detection, which consists of $150$ known intent classes from $10$ domains. The dataset includes $22,500$ in-scope queries and $1,200$ out-of-scope queries. For the in-scope ones, we follow the original splitting, i.e., $15,000$, $3,000$ and $4,500$ for training, validation, and testing respectively. For the out-of-scope ones, we group all of the $1,200$ queries into the test set. 

\textbf{StackOverflow}~\citep{DBLP:conf/naacl/XuWTXZWH15} consists of $20$ classes with $1,000$ examples in each class. We follow the original splitting, i.e., $12,000$ for training, $2,000$ for validation, and $6,000$ for test.

\textbf{Banking}~\citep{DBLP:journals/corr/abs-2003-04807} is a fine-grained intent detection dataset in the banking domain. It consists of $9,003$, $1,000$, and $3,080$ user queries in the training, validation, and test sets respectively.

\textbf{M-CID}~\citep{arora2020cross} is a recently released dataset related to Covid-$19$. We use the English subset of this dataset referred to as M-CID-EN in our experiments, which covers $16$ intent classes. The splitting of M-CID-EN is $1,258$ for training, $148$ for validation, and $339$ for test. 

We extensively compare our method with the following unknown intent detection methods.
\begin{itemize}
\item \textbf{Maximum Softmax Probability (MSP)}~\citep{DBLP:conf/iclr/HendrycksG17} employs the confidence score derived from the maximum softmax probability to predict the class of a sample. The idea under the hood is that the lower the confidence score is, the more likely the sample is of an unknown intent class.

\item \textbf{DOC}~\citep{DBLP:conf/emnlp/ShuXL17} considers to construct $m$ $1$-vs-rest sigmoid classifiers for $m$ seen classes respectively. It uses the maximum probability from these classifiers as the confidence score to conduct classification.

\item \textbf{SEG}~\citep{DBLP:conf/acl/YanFLLZWL20} models the intent distribution as a margin-constrained Gaussian mixture distribution and uses an additional outlier detector -- local outlier factor (LOF)~\citep{10.1145/335191.335388} to achieve unknown intent detection. 

\item \textbf{LMCL}~\citep{DBLP:conf/acl/LinX19} considers to learn discriminative embeddings with a large margin cosine loss. It also uses LOF as the outlier detection algorithm.

\item \textbf{Softmax}~\citep{DBLP:conf/acl/YanFLLZWL20} uses a softmax loss to learn discriminative features based on the training dataset, which also requires an additional outlier detector such as LOF for detecting the unknown intents. 
\end{itemize}

\begin{figure*}[t]
\begin{subfigure}{0.32\textwidth}
  \centering
  \includegraphics[width=1.1\linewidth]{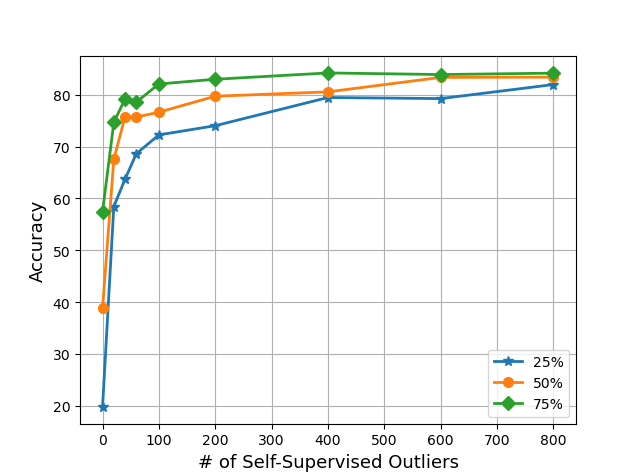}
  \caption{ }
  \label{fig:convex_acc}
\end{subfigure}%
\begin{subfigure}{0.32\textwidth}
  \centering
  \includegraphics[width=1.1\linewidth]{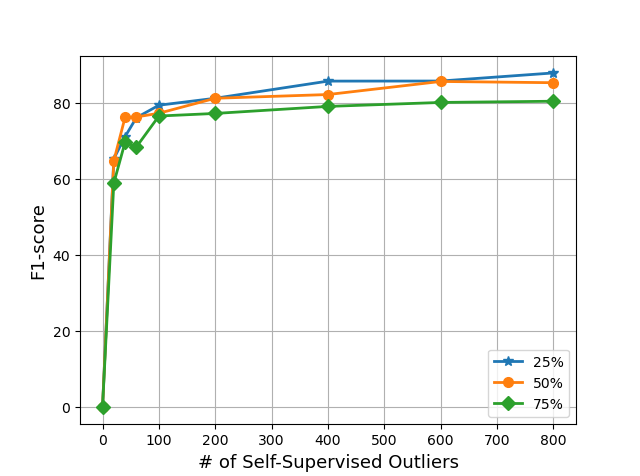}
  \caption{}
  \label{fig:convex_f1_unknown}
\end{subfigure}%
\begin{subfigure}{.32\textwidth}
  \centering
  \includegraphics[width=1.1\linewidth]{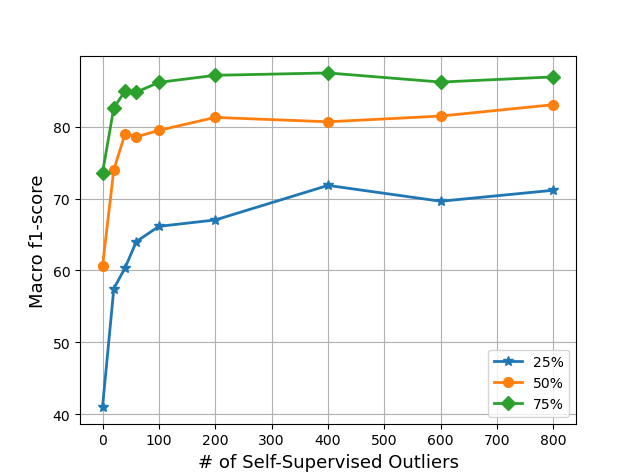}
  \caption{}
  \label{fig:convex_f1}
\end{subfigure}

\begin{subfigure}{.32\textwidth}
  \centering
  \includegraphics[width=1.1\linewidth]{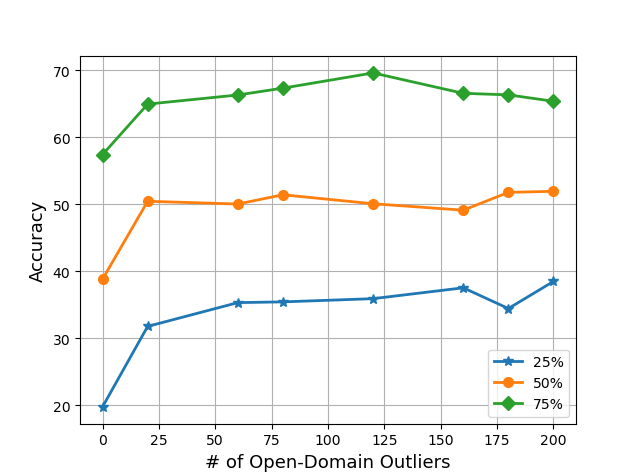}
  \caption{}
  \label{fig:open_domain_acc}
\end{subfigure}%
\begin{subfigure}{.32\textwidth}
  \centering
  \includegraphics[width=1.1\linewidth]{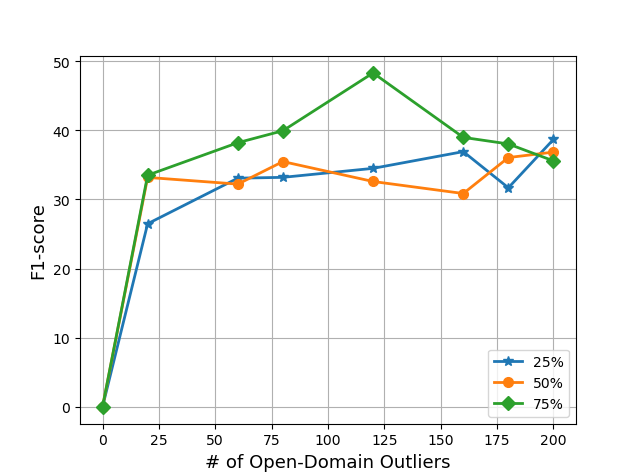}
  \caption{}
  \label{fig:open_domain_f1_unknown}
\end{subfigure}%
\begin{subfigure}{.32\textwidth}
  \centering
  \includegraphics[width=1.1\linewidth]{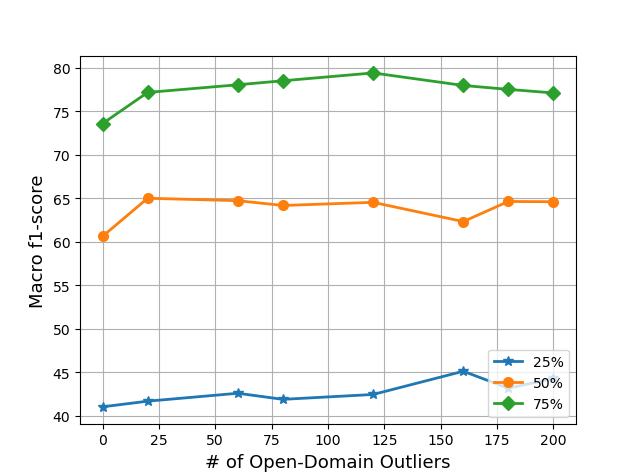}
  \caption{}
  \label{fig:open_domain_f1}
\end{subfigure}
\caption{Effect of the number of pseudo outliers on CLINC150. 
(a), (b), and (c) display overall accuracy, f1-score on the unknown class and overall macro f1-score with varying number of self-supervised outliers respectively. 
(d), (e), and (f) display the corresponding results with varying number of open-domain outliers.}
\label{fig:numbers}
\end{figure*}
\subsection{Experimental Setup and Evaluation Metrics}
To compare with existing methods, we follow the setting in LMCL~\citep{DBLP:conf/acl/LinX19}. Specifically, for each dataset, we randomly sample $75\%$, $50\%$, and $25\%$ of the intent classes from the training set as the known classes to conduct training, and we set aside the rest as the unknown classes for test. 
Notice that for training and validation, we only use data within the chosen known classes and do not expose our model to any of test-time outliers. 
Unless otherwise specified, in each training batch, we keep the ratio of inliers, open-domain outliers and self-supervised outliers roughly as $1:1:4$. This setting is empirically chosen and affected by the memory limit of NVIDIA 2080TI GPU, which we use for conducting the experiments. The number of pseudo outliers can be adjusted according to different environments, and a larger number of self-supervised outliers typically takes more time to converge.

We use Pytorch~\citep{DBLP:conf/nips/PaszkeGMLBCKLGA19} as the backend to conduct the experiments. We use the pretrained BERT mdoel (\emph{bert-base-uncased}) provided by~\citet{wolf2019huggingface} as the encoder for utterances. We use the output vector of the special classification token [CLS] as the utterance embedding and fix its dimension as $768$ by default throughout all of our experiments. To ensure a fair comparison, all baselines and our model use the same encoder. 



For model optimization, we use AdamW provided by \citet{wolf2019huggingface} to fine-tune BERT and Adam proposed by \citet{DBLP:journals/corr/KingmaB14} to train the MLP clasisfier $\Phi(\cdot)$. We set the learning rate for BERT as $1e^{-5}$ as suggested by \citet{DBLP:conf/naacl/DevlinCLT19}. For the MLP clasisfier, the learning rate is fixed as $1e^{-4}$. Notice that the fine-tuning of BERT is conducted simultaneously with the training of the classifier $\Phi(\cdot)$ with the same cross-entropy loss. 
The MLP classifier $\Phi(\cdot)$ has a two-layer architecture with [$1024$, $1024$] as hidden units.
The temperature parameter $\tau$ is selected by cross-validation and set as $0.1$ in all experiments.

Following LMCL~\citep{DBLP:conf/acl/LinX19}, we use overall accuracy and macro f1-score as evaluation metrics. All results reported in this section are the average of $10$ runs with different random seeds, and each run is stopped until reaching a plateau on the validation set. For baselines, we follow their original training settings except using the aforementioned BERT as text encoder.  
\begin{figure}
  \centering
  \includegraphics[width=1.\linewidth,height=0.19\textheight]{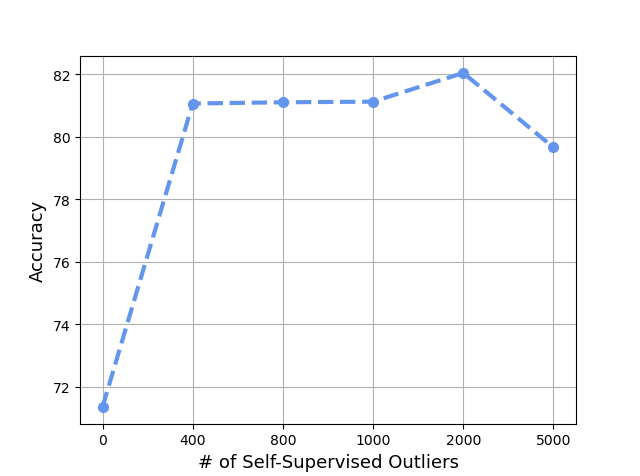}
\caption{Effect of the number of self-supervised outliers on overall intent detection accuracy under the 75\% setting of Banking.}
\label{fig:convex_num}
\end{figure}

\subsection{Result Analysis}
We present our main results in Table~\ref{mainresult1} and Table~\ref{mainresult2}. Specifically, Table~\ref{mainresult1} gives results in overall accuracy and macro f1-score for all classes including the outlier class, while Table~\ref{mainresult2} shows results in macro f1-score for the known classes and f1-score for the outlier class respectively. It can be seen that, on all benchmarks and in almost every setting, our model significantly outperforms the baselines. As shown in Table~\ref{mainresult2}, our method achieves favorable performance on both unknown and known intent classes simultaneously.

It is worth mentioning that the large improvements of our method in scenarios with small labeled training sets (25\% and 50\% settings) indicate its great potential in real-life applications, since a practical dialogue system often needs to deal with a larger proportion of outliers than inliers due to different user demographic, ignorance/unfamiliarity of/with the platform, and limited intent classes recognized by the system (especially at the early development stage). 

More importantly, referring to Table~\ref{mainresult2}, as the proportion of known intents increases, it can be seen that the performance gains of the baselines mainly lie in the known classes. In contrast, our method can strike a better balance between the known and unknown classes without relying on additional outlier detector, margin tuning, and threshold selection, demonstrating its high effectiveness and generality. 
Take the Softmax baseline for example, in the $75\%$ case of CLINC150, it achieves a slightly higher result than our model on the known classes but a substantially lower result on the unknown ones. 

\subsection{Effect of Pseudo Outliers}
We conduct an ablation study on the effectiveness of the two kinds of pseudo outliers and summarize the results in Table~\ref{table:ablation}. The first row of the three settings ($25\%$, $50\%$, and $75\%$) stands for training solely with the labeled examples of CLINC150 without using any pseudo outliers. In general, self-supervised synthetic outliers and open-domain outliers both lead to positive effects on classification performance. For each setting, comparing the second row with the third, we can observe that the synthetic outliers produced by convex combinations lead to a much larger performance gain than that of pre-collected open-domain outliers. Finally, combining them for training leads to the best results, as shown in the fourth row of each setting.

Next, we conduct experiments to study the impact of varying the number of the two kinds of pseudo outliers separately, as shown in Figure~\ref{fig:numbers}. We first fix the number of open-domain outliers as zero and then increase the number of self-supervised outliers. The results are displayed in Figure~\ref{fig:numbers} (a), (b) and (c). In particular, as the number of self-supervised outliers grows, the performance first increases quickly and then grows slowly. On the other hand, we fix the number of self-supervised outliers as zero and then increases the number of open-domain outliers. The results are shown in Figure~\ref{fig:numbers}  (d), (e) and (f), where it can be seen that dozens of open-domain outliers already can bring significant improvements, though the gain is much smaller compared to that of the self-supervised outliers.

Finally, we investigate the impact of the number of self-supervised outliers on overall intent detection accuracy with both the number of inliers and the number of open-domain outliers fixed as 100 per training batch. 
As shown in Figure~\ref{fig:convex_num}, we increase the number of self-supervised outliers from 0 to $5000$. Note that $400$ is the default setting used in Table~\ref{mainresult1} and Table~\ref{mainresult2}. We can see that comparable results can be obtained for a wide range of numbers. However, when the number grows to $5000$, the performance exhibits a significant drop. We hypothesize that as the number increases, the generated synthetic outliers may be less accurate, because some convex combinations may fall within the scope of known classes.


To summarize, self-supervised outliers play a much more important role than open-domain outliers for unknown intent classification. Self-supervised outliers not only provide better learning signals for the unknown intents, but also impose an important positive effect on the known ones. For the open-domain outliers, if used alone, they can only provide limited benefit. But in combination with the self-supervised ones, they can further enhance the performance.
\begin{table}[t]
\centering
\scalebox{0.8}{
\begin{tabular}{lccccc}
\specialrule{0.15em}{0.2em}{0.2em}
&$\mathcal{D}_{co}^{tr}$&$\mathcal{D}_{fo}^{tr}$&Acc&Macro-F1 & F1 Unknown\\
\specialrule{0.05em}{0.1em}{0.2em}
\multirow{4}{*}{25\%} 
                    & & &19.79&41.05&-\\
                    &\checkmark & &81.96&71.15&87.8\\
                    & & \checkmark&37.55&45.14&36.91\\
                    &\checkmark & \checkmark&88.44&80.73&92.35\\
\specialrule{0.05em}{0.1em}{0.2em}
\multirow{4}{*}{50\%} 
                    & & &38.78&60.35&-\\
                    &\checkmark & &83.12&82.62&85.03\\
                    & & \checkmark&48.62&63.19&28.82\\
                    &\checkmark & \checkmark&88.33&86.67&90.30\\
\specialrule{0.05em}{0.1em}{0.2em}
\multirow{4}{*}{75\%} 
                    & & &57.43&73.6&-\\
                    &\checkmark & &84.16&86.9&80.36\\
                    & & \checkmark&69.61&79.42&48.29\\
                    &\checkmark & \checkmark&88.08&89.43&86.28\\
\specialrule{0.15em}{0.2em}{0.2em}
\end{tabular}
}
\caption{\label{table:ablation}
An ablation study on the effectiveness of pseudo outliers.
}
\end{table}

\begin{table}
\centering
\scalebox{1.}{
\begin{tabular}{lccc}
\specialrule{0.1em}{0.2em}{0.2em}
&$\mathcal{D}_{fo}^{tr}$&Acc&Macro-F1\\
\specialrule{0.05em}{0.1em}{0.2em}
\multirow{3}{*}{25\%} 
                   &Open-bank  & 89.36&81.22\\
                   & Open-stack &88.38 &80.42\\
                   & Open-big &88.44 &80.73\\
\specialrule{0.05em}{0.1em}{0.2em}
\multirow{3}{*}{50\%} 
                    & Open-bank&87.35 &86.41\\
                    &Open-stack&88.23 &86.37\\
                    &Open-big & 88.33&86.67\\
\specialrule{0.05em}{0.1em}{0.2em}
\multirow{3}{*}{75\%} 
                    & Open-bank& 87.19&89.33\\
                    & Open-stack&87.52 &89.17\\
                    &Open-big & 88.08&89.43\\
\specialrule{0.1em}{0.2em}{0.2em}
\end{tabular}
}

\caption{
 Results on CLINC150 with different sets of open-domain outliers.
 }
 \label{table:ablation_open_domain_outlier}
\end{table}
\subsection{Selection of Open-Domain Outliers}
To demonstrate the flexibility of our method in selecting open-domain outliers as described in Section~\ref{subsec: building outlier}, we train our model on CLINC150 using open-domain outliers from different sources. The results are summarized in Table~\ref{table:ablation_open_domain_outlier}. Specifically, Open-bank and Open-stack stand for using the training set of Banking and StackOverflow as the source of open-domain outliers respectively. Open-big stands for the source of open-domain outliers used in other experiments, which consists of $\sim0.5$ million sentences randomly selected from SQuaD 2.0~\citep{DBLP:conf/acl/RajpurkarJL18}, Yelp~\citep{DBLP:conf/cikm/MengSZ018}, and IMDB~\citep{DBLP:conf/acl/MaasDPHNP11}. It can be seen that the performance of our model is insensitive to the selection of open-domain outliers. 

\begin{figure}
\begin{subfigure}{0.23\textwidth}
  \centering
  \includegraphics[width=1.\linewidth,height=0.11\textheight]{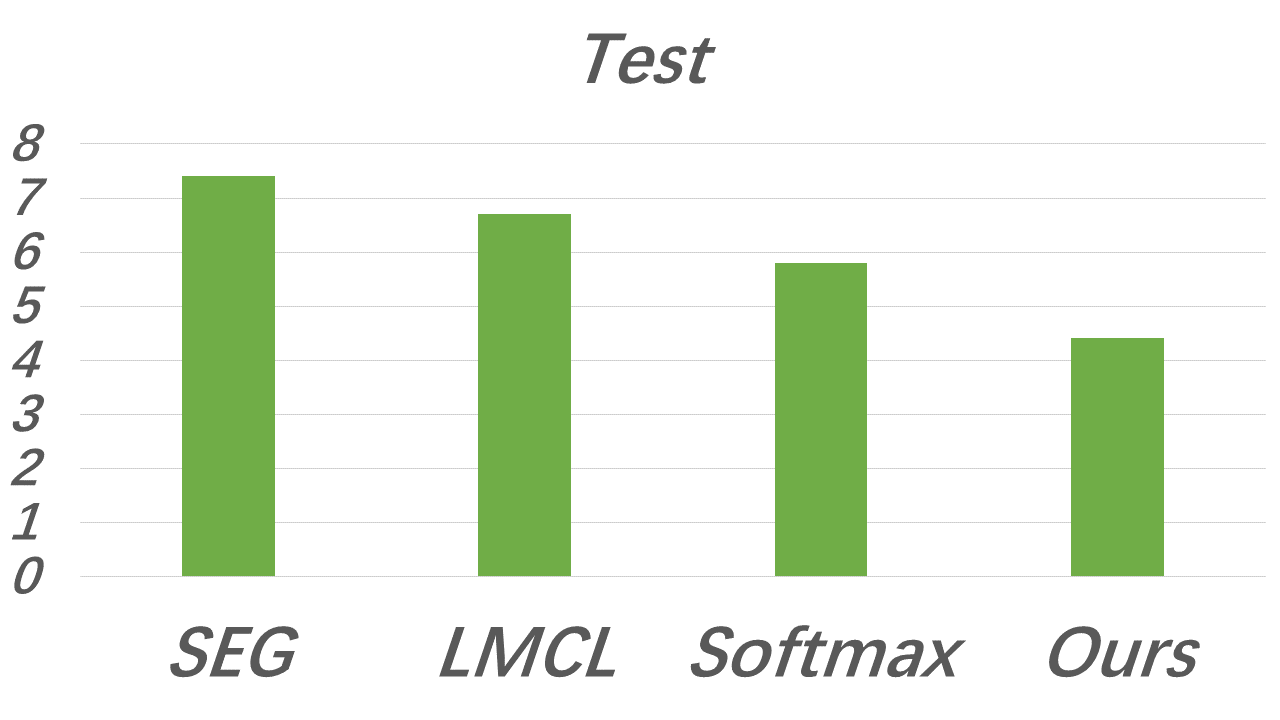}
\end{subfigure}%
\begin{subfigure}{0.23\textwidth}
  \centering
  \includegraphics[width=1.\linewidth,height=0.11\textheight]{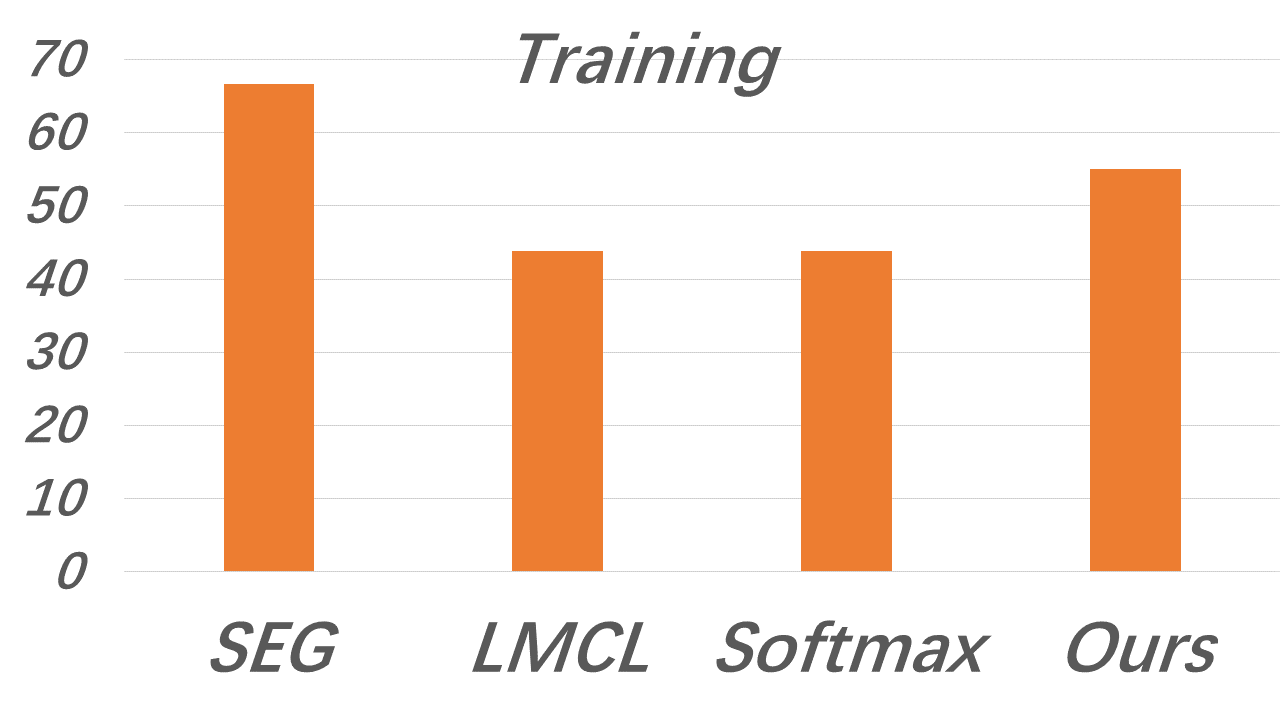}
\end{subfigure}
\caption{Comparison of training time (per epoch) and test time with baselines.}
\label{fig:efficiency}
\end{figure}

\subsection{Efficiency}
We provide a quantitative comparison on the training and test efficiency for our method and the baselines, by calculating the average time (in seconds) for training per epoch and the total time for testing under the $75\%$ setting. Here, we only compare with the strongest baselines. As shown in Figure~\ref{fig:efficiency}, even with the pseudo outliers, the training time of our method is comparable to that of the baselines. Importantly, in the test stage, our method demonstrates significant advantages in efficiency, which needs much less time to predict intent classes for all samples in the test set.


\section{Conclusion}

We have proposed a simple, effective, and efficient approach for out-of-scope intent detection by overcoming the limitation of previous methods via matching train-test conditions. Particularly, at the training stage, we construct self-supervised and open-domain outliers to improve model generalization and simulate real outliers in the test stage. Extensive experiments on four dialogue datasets show that our approach significantly outperforms state-of-the-art methods. In the future, we plan to investigate the theoretical underpinnings of our approach and apply it to more applications.



\section*{Acknowledgments}

We would like to thank the anonymous reviewers for their helpful comments. This research was supported by the grant HK ITF UIM/377.

\bibliographystyle{acl_natbib}
\bibliography{anthology,acl2021}

\begin{thebibliography}{42}
\expandafter\ifx\csname natexlab\endcsname\relax\def\natexlab#1{#1}\fi

\bibitem[{Arora et~al.(2020)Arora, Shrivastava, Mohit, Lecanda, and
  Aly}]{arora2020cross}
Abhinav Arora, Akshat Shrivastava, Mrinal Mohit, Lorena Sainz-Maza Lecanda, and
  Ahmed Aly. 2020.
\newblock Cross-lingual transfer learning for intent detection of covid-19
  utterances.

\bibitem[{Breunig et~al.(2000)Breunig, Kriegel, Ng, and
  Sander}]{10.1145/335191.335388}
Markus~M. Breunig, Hans-Peter Kriegel, Raymond~T. Ng, and J\"{o}rg Sander.
  2000.
\newblock \href {https://doi.org/10.1145/335191.335388} {Lof: Identifying
  density-based local outliers}.
\newblock \emph{SIGMOD Rec.}, 29(2):93–104.

\bibitem[{Casanueva et~al.(2020)Casanueva, Temcinas, Gerz, Henderson, and
  Vulic}]{DBLP:journals/corr/abs-2003-04807}
I{\~{n}}igo Casanueva, Tadas Temcinas, Daniela Gerz, Matthew Henderson, and
  Ivan Vulic. 2020.
\newblock \href {http://arxiv.org/abs/2003.04807} {Efficient intent detection
  with dual sentence encoders}.
\newblock \emph{CoRR}, abs/2003.04807.

\bibitem[{Chandola et~al.(2009)Chandola, Banerjee, and
  Kumar}]{chandola2009anomaly}
Varun Chandola, Arindam Banerjee, and Vipin Kumar. 2009.
\newblock Anomaly detection: A survey.
\newblock \emph{ACM computing surveys (CSUR)}, 41(3):1--58.

\bibitem[{Cortes and Vapnik(1995)}]{cortes1995support}
Corinna Cortes and Vladimir Vapnik. 1995.
\newblock Support-vector networks.
\newblock \emph{Machine learning}, 20(3):273--297.

\bibitem[{Devlin et~al.(2019)Devlin, Chang, Lee, and
  Toutanova}]{DBLP:conf/naacl/DevlinCLT19}
Jacob Devlin, Ming{-}Wei Chang, Kenton Lee, and Kristina Toutanova. 2019.
\newblock \href {https://doi.org/10.18653/v1/n19-1423} {{BERT:} pre-training of
  deep bidirectional transformers for language understanding}.
\newblock In \emph{Proceedings of the 2019 Conference of the North American
  Chapter of the Association for Computational Linguistics: Human Language
  Technologies, {NAACL-HLT} 2019, Minneapolis, MN, USA, June 2-7, 2019, Volume
  1 (Long and Short Papers)}, pages 4171--4186. Association for Computational
  Linguistics.

\bibitem[{Goodfellow et~al.(2014)Goodfellow, Pouget-Abadie, Mirza, Xu,
  Warde-Farley, Ozair, Courville, and Bengio}]{goodfellow2014generative}
Ian~J Goodfellow, Jean Pouget-Abadie, Mehdi Mirza, Bing Xu, David Warde-Farley,
  Sherjil Ozair, Aaron Courville, and Yoshua Bengio. 2014.
\newblock Generative adversarial networks.
\newblock \emph{arXiv preprint arXiv:1406.2661}.

\bibitem[{Goodfellow et~al.(2015)Goodfellow, Shlens, and
  Szegedy}]{DBLP:journals/corr/GoodfellowSS14}
Ian~J. Goodfellow, Jonathon Shlens, and Christian Szegedy. 2015.
\newblock \href {http://arxiv.org/abs/1412.6572} {Explaining and harnessing
  adversarial examples}.
\newblock In \emph{3rd International Conference on Learning Representations,
  {ICLR} 2015, San Diego, CA, USA, May 7-9, 2015, Conference Track
  Proceedings}.

\bibitem[{Hendrycks et~al.(2020{\natexlab{a}})Hendrycks, Basart, Mu, Kadavath,
  Wang, Dorundo, Desai, Zhu, Parajuli, Guo et~al.}]{hendrycks2020many}
Dan Hendrycks, Steven Basart, Norman Mu, Saurav Kadavath, Frank Wang, Evan
  Dorundo, Rahul Desai, Tyler Zhu, Samyak Parajuli, Mike Guo, et~al.
  2020{\natexlab{a}}.
\newblock The many faces of robustness: A critical analysis of
  out-of-distribution generalization.
\newblock \emph{arXiv preprint arXiv:2006.16241}.

\bibitem[{Hendrycks and Gimpel(2017)}]{DBLP:conf/iclr/HendrycksG17}
Dan Hendrycks and Kevin Gimpel. 2017.
\newblock \href {https://openreview.net/forum?id=Hkg4TI9xl} {A baseline for
  detecting misclassified and out-of-distribution examples in neural networks}.
\newblock In \emph{5th International Conference on Learning Representations,
  {ICLR} 2017, Toulon, France, April 24-26, 2017, Conference Track
  Proceedings}. OpenReview.net.

\bibitem[{Hendrycks et~al.(2019)Hendrycks, Lee, and
  Mazeika}]{DBLP:conf/icml/HendrycksLM19}
Dan Hendrycks, Kimin Lee, and Mantas Mazeika. 2019.
\newblock \href {http://proceedings.mlr.press/v97/hendrycks19a.html} {Using
  pre-training can improve model robustness and uncertainty}.
\newblock In \emph{Proceedings of the 36th International Conference on Machine
  Learning, {ICML} 2019, 9-15 June 2019, Long Beach, California, {USA}},
  volume~97 of \emph{Proceedings of Machine Learning Research}, pages
  2712--2721. {PMLR}.

\bibitem[{Hendrycks et~al.(2020{\natexlab{b}})Hendrycks, Liu, Wallace,
  Dziedzic, Krishnan, and Song}]{DBLP:conf/acl/HendrycksLWDKS20}
Dan Hendrycks, Xiaoyuan Liu, Eric Wallace, Adam Dziedzic, Rishabh Krishnan, and
  Dawn Song. 2020{\natexlab{b}}.
\newblock \href {https://doi.org/10.18653/v1/2020.acl-main.244} {Pretrained
  transformers improve out-of-distribution robustness}.
\newblock In \emph{Proceedings of the 58th Annual Meeting of the Association
  for Computational Linguistics, {ACL} 2020, Online, July 5-10, 2020}, pages
  2744--2751. Association for Computational Linguistics.

\bibitem[{Hendrycks et~al.(2018)Hendrycks, Mazeika, and
  Dietterich}]{hendrycks2018deep}
Dan Hendrycks, Mantas Mazeika, and Thomas Dietterich. 2018.
\newblock Deep anomaly detection with outlier exposure.
\newblock \emph{arXiv preprint arXiv:1812.04606}.

\bibitem[{Hodge and Austin(2004)}]{DBLP:journals/air/HodgeA04}
Victoria~J. Hodge and Jim Austin. 2004.
\newblock \href {https://doi.org/10.1023/B:AIRE.0000045502.10941.a9} {A survey
  of outlier detection methodologies}.
\newblock \emph{Artif. Intell. Rev.}, 22(2):85--126.

\bibitem[{Kingma and Ba(2015)}]{DBLP:journals/corr/KingmaB14}
Diederik~P. Kingma and Jimmy Ba. 2015.
\newblock \href {http://arxiv.org/abs/1412.6980} {Adam: {A} method for
  stochastic optimization}.
\newblock In \emph{3rd International Conference on Learning Representations,
  {ICLR} 2015, San Diego, CA, USA, May 7-9, 2015, Conference Track
  Proceedings}.

\bibitem[{Larson et~al.(2019)Larson, Mahendran, Peper, Clarke, Lee, Hill,
  Kummerfeld, Leach, Laurenzano, Tang et~al.}]{larson2019evaluation}
Stefan Larson, Anish Mahendran, Joseph~J Peper, Christopher Clarke, Andrew Lee,
  Parker Hill, Jonathan~K Kummerfeld, Kevin Leach, Michael~A Laurenzano,
  Lingjia Tang, et~al. 2019.
\newblock An evaluation dataset for intent classification and out-of-scope
  prediction.
\newblock In \emph{Proceedings of the 2019 Conference on Empirical Methods in
  Natural Language Processing and the 9th International Joint Conference on
  Natural Language Processing (EMNLP-IJCNLP)}, pages 1311--1316.

\bibitem[{Lee et~al.(2017)Lee, Lee, Lee, and Shin}]{lee2017training}
Kimin Lee, Honglak Lee, Kibok Lee, and Jinwoo Shin. 2017.
\newblock Training confidence-calibrated classifiers for detecting
  out-of-distribution samples.
\newblock \emph{arXiv preprint arXiv:1711.09325}.

\bibitem[{Liang et~al.(2018)Liang, Li, and
  Srikant}]{dcbe7abf4db64d1b89bf9802585660ed}
Shiyu Liang, Yixuan Li, and R.~Srikant. 2018.
\newblock Enhancing the reliability of out-of-distribution image detection in
  neural networks.
\newblock 6th International Conference on Learning Representations, ICLR 2018 ;
  Conference date: 30-04-2018 Through 03-05-2018.

\bibitem[{Lin and Xu(2019)}]{DBLP:conf/acl/LinX19}
Ting{-}En Lin and Hua Xu. 2019.
\newblock \href {https://doi.org/10.18653/v1/p19-1548} {Deep unknown intent
  detection with margin loss}.
\newblock In \emph{Proceedings of the 57th Conference of the Association for
  Computational Linguistics, {ACL} 2019, Florence, Italy, July 28- August 2,
  2019, Volume 1: Long Papers}, pages 5491--5496. Association for Computational
  Linguistics.

\bibitem[{Liu et~al.(2008)Liu, Ting, and Zhou}]{10.1109/ICDM.2008.17}
Fei~Tony Liu, Kai~Ming Ting, and Zhi-Hua Zhou. 2008.
\newblock \href {https://doi.org/10.1109/ICDM.2008.17} {Isolation forest}.
\newblock In \emph{Proceedings of the 2008 Eighth IEEE International Conference
  on Data Mining}, ICDM '08, page 413–422, USA. IEEE Computer Society.

\bibitem[{Maas et~al.(2011)Maas, Daly, Pham, Huang, Ng, and
  Potts}]{DBLP:conf/acl/MaasDPHNP11}
Andrew~L. Maas, Raymond~E. Daly, Peter~T. Pham, Dan Huang, Andrew~Y. Ng, and
  Christopher Potts. 2011.
\newblock \href {https://www.aclweb.org/anthology/P11-1015/} {Learning word
  vectors for sentiment analysis}.
\newblock In \emph{The 49th Annual Meeting of the Association for Computational
  Linguistics: Human Language Technologies, Proceedings of the Conference,
  19-24 June, 2011, Portland, Oregon, {USA}}, pages 142--150. The Association
  for Computer Linguistics.

\bibitem[{Meng et~al.(2018)Meng, Shen, Zhang, and
  Han}]{DBLP:conf/cikm/MengSZ018}
Yu~Meng, Jiaming Shen, Chao Zhang, and Jiawei Han. 2018.
\newblock \href {https://doi.org/10.1145/3269206.3271737} {Weakly-supervised
  neural text classification}.
\newblock In \emph{Proceedings of the 27th {ACM} International Conference on
  Information and Knowledge Management, {CIKM} 2018, Torino, Italy, October
  22-26, 2018}, pages 983--992. {ACM}.

\bibitem[{Orhan(2019)}]{DBLP:journals/corr/abs-1907-07640}
A.~Emin Orhan. 2019.
\newblock \href {http://arxiv.org/abs/1907.07640} {Robustness properties of
  facebook's resnext {WSL} models}.
\newblock \emph{CoRR}, abs/1907.07640.

\bibitem[{Paszke et~al.(2019)Paszke, Gross, Massa, Lerer, Bradbury, Chanan,
  Killeen, Lin, Gimelshein, Antiga, Desmaison, K{\"{o}}pf, Yang, DeVito,
  Raison, Tejani, Chilamkurthy, Steiner, Fang, Bai, and
  Chintala}]{DBLP:conf/nips/PaszkeGMLBCKLGA19}
Adam Paszke, Sam Gross, Francisco Massa, Adam Lerer, James Bradbury, Gregory
  Chanan, Trevor Killeen, Zeming Lin, Natalia Gimelshein, Luca Antiga, Alban
  Desmaison, Andreas K{\"{o}}pf, Edward Yang, Zachary DeVito, Martin Raison,
  Alykhan Tejani, Sasank Chilamkurthy, Benoit Steiner, Lu~Fang, Junjie Bai, and
  Soumith Chintala. 2019.
\newblock \href
  {https://proceedings.neurips.cc/paper/2019/hash/bdbca288fee7f92f2bfa9f7012727740-Abstract.html}
  {Pytorch: An imperative style, high-performance deep learning library}.
\newblock In \emph{Advances in Neural Information Processing Systems 32: Annual
  Conference on Neural Information Processing Systems 2019, NeurIPS 2019,
  December 8-14, 2019, Vancouver, BC, Canada}, pages 8024--8035.

\bibitem[{Rajpurkar et~al.(2018)Rajpurkar, Jia, and
  Liang}]{DBLP:conf/acl/RajpurkarJL18}
Pranav Rajpurkar, Robin Jia, and Percy Liang. 2018.
\newblock \href {https://doi.org/10.18653/v1/P18-2124} {Know what you don't
  know: Unanswerable questions for squad}.
\newblock In \emph{Proceedings of the 56th Annual Meeting of the Association
  for Computational Linguistics, {ACL} 2018, Melbourne, Australia, July 15-20,
  2018, Volume 2: Short Papers}, pages 784--789. Association for Computational
  Linguistics.

\bibitem[{Rousseeuw and Driessen(1999)}]{article}
Peter Rousseeuw and Katrien Driessen. 1999.
\newblock \href {https://doi.org/10.1080/00401706.1999.10485670} {A fast
  algorithm for the minimum covariance determinant estimator}.
\newblock \emph{Technometrics}, 41:212--223.

\bibitem[{Ruff et~al.(2021)Ruff, Kauffmann, Vandermeulen, Montavon, Samek,
  Kloft, Dietterich, and M{\"{u}}ller}]{DBLP:journals/pieee/RuffKVMSKDM21}
Lukas Ruff, Jacob~R. Kauffmann, Robert~A. Vandermeulen, Gr{\'{e}}goire
  Montavon, Wojciech Samek, Marius Kloft, Thomas~G. Dietterich, and
  Klaus{-}Robert M{\"{u}}ller. 2021.
\newblock \href {https://doi.org/10.1109/JPROC.2021.3052449} {A unifying review
  of deep and shallow anomaly detection}.
\newblock \emph{Proc. {IEEE}}, 109(5):756--795.

\bibitem[{Ryu et~al.(2017)Ryu, Kim, Choi, Yu, and
  Lee}]{10.1016/j.patrec.2017.01.008}
Seonghan Ryu, Seokhwan Kim, Junhwi Choi, Hwanjo Yu, and Gary~Geunbae Lee. 2017.
\newblock \href {https://doi.org/10.1016/j.patrec.2017.01.008} {Neural sentence
  embedding using only in-domain sentences for out-of-domain sentence detection
  in dialog systems}.
\newblock \emph{Pattern Recogn. Lett.}, 88(C):26–32.

\bibitem[{Ryu et~al.(2018)Ryu, Koo, Yu, and Lee}]{ryu-etal-2018-domain}
Seonghan Ryu, Sangjun Koo, Hwanjo Yu, and Gary~Geunbae Lee. 2018.
\newblock \href {https://doi.org/10.18653/v1/D18-1077} {Out-of-domain detection
  based on generative adversarial network}.
\newblock In \emph{Proceedings of the 2018 Conference on Empirical Methods in
  Natural Language Processing}, pages 714--718, Brussels, Belgium. Association
  for Computational Linguistics.

\bibitem[{{Schölkopf} et~al.(2001){Schölkopf}, {Platt}, {Shawe-Taylor},
  {Smola}, and {Williamson}}]{6790022}
B.~{Schölkopf}, J.~C. {Platt}, J.~{Shawe-Taylor}, A.~J. {Smola}, and R.~C.
  {Williamson}. 2001.
\newblock \href {https://doi.org/10.1162/089976601750264965} {Estimating the
  support of a high-dimensional distribution}.
\newblock \emph{Neural Computation}, 13(7):1443--1471.

\bibitem[{Shu et~al.(2017)Shu, Xu, and Liu}]{DBLP:conf/emnlp/ShuXL17}
Lei Shu, Hu~Xu, and Bing Liu. 2017.
\newblock \href {https://doi.org/10.18653/v1/d17-1314} {{DOC:} deep open
  classification of text documents}.
\newblock In \emph{Proceedings of the 2017 Conference on Empirical Methods in
  Natural Language Processing, {EMNLP} 2017, Copenhagen, Denmark, September
  9-11, 2017}, pages 2911--2916. Association for Computational Linguistics.

\bibitem[{Tan et~al.(2019)Tan, Yu, Wang, Wang, Potdar, Chang, and
  Yu}]{tan-etal-2019-domain}
Ming Tan, Yang Yu, Haoyu Wang, Dakuo Wang, Saloni Potdar, Shiyu Chang, and
  Mo~Yu. 2019.
\newblock \href {https://doi.org/10.18653/v1/D19-1364} {Out-of-domain detection
  for low-resource text classification tasks}.
\newblock In \emph{Proceedings of the 2019 Conference on Empirical Methods in
  Natural Language Processing and the 9th International Joint Conference on
  Natural Language Processing (EMNLP-IJCNLP)}, pages 3566--3572, Hong Kong,
  China. Association for Computational Linguistics.

\bibitem[{Thulasidasan et~al.(2019)Thulasidasan, Chennupati, Bilmes,
  Bhattacharya, and Michalak}]{DBLP:conf/nips/ThulasidasanCBB19}
Sunil Thulasidasan, Gopinath Chennupati, Jeff~A. Bilmes, Tanmoy Bhattacharya,
  and Sarah Michalak. 2019.
\newblock \href
  {https://proceedings.neurips.cc/paper/2019/hash/36ad8b5f42db492827016448975cc22d-Abstract.html}
  {On mixup training: Improved calibration and predictive uncertainty for deep
  neural networks}.
\newblock In \emph{Advances in Neural Information Processing Systems 32: Annual
  Conference on Neural Information Processing Systems 2019, NeurIPS 2019,
  December 8-14, 2019, Vancouver, BC, Canada}, pages 13888--13899.

\bibitem[{Vinyals et~al.(2016)Vinyals, Blundell, Lillicrap, Kavukcuoglu, and
  Wierstra}]{vinyals2016matching}
Oriol Vinyals, Charles Blundell, Timothy Lillicrap, Koray Kavukcuoglu, and Daan
  Wierstra. 2016.
\newblock Matching networks for one shot learning.
\newblock In \emph{Proceedings of the 30th International Conference on Neural
  Information Processing Systems}, pages 3637--3645.

\bibitem[{Wolf et~al.(2019)Wolf, Debut, Sanh, Chaumond, Delangue, Moi, Cistac,
  Rault, Louf, Funtowicz et~al.}]{wolf2019huggingface}
Thomas Wolf, Lysandre Debut, Victor Sanh, Julien Chaumond, Clement Delangue,
  Anthony Moi, Pierric Cistac, Tim Rault, R{\'e}mi Louf, Morgan Funtowicz,
  et~al. 2019.
\newblock Huggingface's transformers: State-of-the-art natural language
  processing.
\newblock \emph{arXiv preprint arXiv:1910.03771}.

\bibitem[{Xu et~al.(2015)Xu, Wang, Tian, Xu, Zhao, Wang, and
  Hao}]{DBLP:conf/naacl/XuWTXZWH15}
Jiaming Xu, Peng Wang, Guanhua Tian, Bo~Xu, Jun Zhao, Fangyuan Wang, and
  Hongwei Hao. 2015.
\newblock \href {https://doi.org/10.3115/v1/w15-1509} {Short text clustering
  via convolutional neural networks}.
\newblock In \emph{Proceedings of the 1st Workshop on Vector Space Modeling for
  Natural Language Processing, VS@NAACL-HLT 2015, June 5, 2015, Denver,
  Colorado, {USA}}, pages 62--69. The Association for Computational
  Linguistics.

\bibitem[{Yan et~al.(2020{\natexlab{a}})Yan, Fan, Li, Liu, Zhang, Wu, and
  Lam}]{DBLP:conf/acl/YanFLLZWL20}
Guangfeng Yan, Lu~Fan, Qimai Li, Han Liu, Xiaotong Zhang, Xiao{-}Ming Wu, and
  Albert Y.~S. Lam. 2020{\natexlab{a}}.
\newblock \href {https://doi.org/10.18653/v1/2020.acl-main.99} {Unknown intent
  detection using gaussian mixture model with an application to zero-shot
  intent classification}.
\newblock In \emph{Proceedings of the 58th Annual Meeting of the Association
  for Computational Linguistics, {ACL} 2020, Online, July 5-10, 2020}, pages
  1050--1060. Association for Computational Linguistics.

\bibitem[{Yan et~al.(2020{\natexlab{b}})Yan, Fan, Li, Liu, Zhang, Wu, and
  Lam}]{yan-etal-2020-unknown}
Guangfeng Yan, Lu~Fan, Qimai Li, Han Liu, Xiaotong Zhang, Xiao-Ming Wu, and
  Albert~Y.S. Lam. 2020{\natexlab{b}}.
\newblock \href {https://doi.org/10.18653/v1/2020.acl-main.99} {Unknown intent
  detection using {G}aussian mixture model with an application to zero-shot
  intent classification}.
\newblock In \emph{Proceedings of the 58th Annual Meeting of the Association
  for Computational Linguistics}, pages 1050--1060, Online. Association for
  Computational Linguistics.

\bibitem[{Zhang et~al.(2021)Zhang, Xu, and Lin}]{Zhang_Xu_Lin_2021}
Hanlei Zhang, Hua Xu, and Ting-En Lin. 2021.
\newblock Deep open intent classification with adaptive decision boundary.
\newblock \emph{Proceedings of the AAAI Conference on Artificial Intelligence},
  35(16):14374--14382.

\bibitem[{Zhang et~al.(2018)Zhang, Ciss{\'{e}}, Dauphin, and
  Lopez{-}Paz}]{DBLP:conf/iclr/ZhangCDL18}
Hongyi Zhang, Moustapha Ciss{\'{e}}, Yann~N. Dauphin, and David Lopez{-}Paz.
  2018.
\newblock \href {https://openreview.net/forum?id=r1Ddp1-Rb} {mixup: Beyond
  empirical risk minimization}.
\newblock In \emph{6th International Conference on Learning Representations,
  {ICLR} 2018, Vancouver, BC, Canada, April 30 - May 3, 2018, Conference Track
  Proceedings}. OpenReview.net.

\bibitem[{Zhang et~al.(2020)Zhang, Hashimoto, Liu, Wu, Wan, Yu, Socher, and
  Xiong}]{zhang-etal-2020-discriminative}
Jianguo Zhang, Kazuma Hashimoto, Wenhao Liu, Chien-Sheng Wu, Yao Wan, Philip
  Yu, Richard Socher, and Caiming Xiong. 2020.
\newblock \href {https://doi.org/10.18653/v1/2020.emnlp-main.411}
  {Discriminative nearest neighbor few-shot intent detection by transferring
  natural language inference}.
\newblock In \emph{Proceedings of the 2020 Conference on Empirical Methods in
  Natural Language Processing (EMNLP)}, pages 5064--5082, Online. Association
  for Computational Linguistics.

\bibitem[{Zimek et~al.(2012)Zimek, Schubert, and Kriegel}]{zimek2012survey}
Arthur Zimek, Erich Schubert, and Hans-Peter Kriegel. 2012.
\newblock A survey on unsupervised outlier detection in high-dimensional
  numerical data.
\newblock \emph{Statistical Analysis and Data Mining: The ASA Data Science
  Journal}, 5(5):363--387.

\end{thebibliography}


\end{document}